\documentclass[11pt,a4paper]{article}
\usepackage[hyperref]{acl2020}
\usepackage{times}
\usepackage{amsmath}
\usepackage{latexsym}
\usepackage{microtype}
\usepackage{graphicx}
\usepackage{array}
\usepackage{xcolor}
\usepackage{floatrow}
\usepackage{booktabs}
\usepackage{caption}
\usepackage{subcaption}

\aclfinalcopy
\title{BERTVision - A Parameter-Efficient Approach for Question Answering}
\author{Siduo Jiang \\
  Microsoft \\
  \small{\texttt{stonejiang@microsoft.com}}
  \And
  Cristopher Benge \\
  Microsoft \\
  \small{\texttt{crbenge@microsoft.com}}
  \And
  William Casey King \\
  Yale University\\
  \small{\texttt{casey.king@yale.edu}}
} 
\date{}

\newcolumntype{L}[1]{>{\raggedright\let\newline\\\arraybackslash\hspace{0pt}}m{#1}}
\newcolumntype{C}[1]{>{\centering\let\newline\\\arraybackslash\hspace{0pt}}m{#1}}
\newcolumntype{R}[1]{>{\raggedleft\let\newline\\\arraybackslash\hspace{0pt}}m{#1}}

\definecolor{laplane}{HTML}{00a598}
\definecolor{berkeleyblue}{HTML}{003262}

\begin{document}

	\maketitle
	\begin{abstract}
We present a highly parameter-efficient approach for Question Answering (QA) that significantly reduces the need for extended BERT fine-tuning. Our method uses information from the hidden state activations of each BERT transformer layer, which is discarded during typical BERT inference. Our best model achieves maximal BERT performance at a fraction of the training time and GPU/TPU expense. Performance is further improved by ensembling our model with BERT’s predictions. Furthermore, we find that near optimal performance can be achieved for QA span annotation using less training data. Our experiments show that this approach works well not only for span annotation, but also for classification, suggesting that it may be extensible to a wider range of tasks.
\footnote{\textit{BERTVision} - so named for our method of peering within BERT for the signal hidden therein.}
\footnote{See GitHub repository: \href{https://github.com/cbenge509/BERTVision}{BERTVision}}
\end{abstract}
	\section{Introduction}

% Tenney1 = tenney-etal-2019-bert
% Tenney2 = DBLP:journals/corr/abs-1905-06316
% Zhu = Zhu2020IncorporatingBI
% Chen = Chen_2020
% Aken = van_Aken_2019
% Vaswani = Vaswani2017
% Devlin = Devlin2019
% distil = sanh2019distilbert
% BLEU = Papineni02bleu:a
% Ma = ma2019universal

The introduction of Transformers \cite{Vaswani2017} has significantly advanced the state-of-the-art for many NLP tasks. The most well-known Transformer-based model is BERT \cite{Devlin2019}. The standard way to use BERT on a specific task is to first download pre-trained weights for the model, then fine-tune these weights on a supervised dataset. However, this procedure can be quite slow, and at times prohibitive for those without a powerful GPU/TPU, or those with limited CPU capacity. Smaller Transformers, such as DistilBERT \cite{sanh2019distilbert}, can fine-tune up to 60\% faster. However, such models tend to consistently underperform full-size BERT on a wide range of tasks. A method that reduces fine-tuning but maintains the same or better performance would make BERT more accessible for practical applications.

To develop our method, we drew inspiration from previous works that use BERT for feature extraction rather than for fine-tuning \citep{Zhu2020IncorporatingBI, Chen_2020}. For example, Zhu et al. showed that the sequence outputs from the final BERT layer can be used as contextualized embeddings to supplement the self-attention mechanism in an encoder/decoder neural translation model. This led to an improvement over the standard Transformer model in all tested language pairs on standard metrics (BLEU score \cite{Papineni02bleu:a}).

One characteristic these studies share with typical BERT inference is that only information from the final layer of BERT is used. However, a study by \cite{tenney-etal-2019-bert} suggests that all layers of BERT carry unique information. By training a series of classifiers within the edge probing framework \cite{DBLP:journals/corr/abs-1905-06316}, the authors computed how much each layer changes performance on eight labeling tasks, and the expected layer at which the model predicts the correct labels. The discovery is that syntactic information is encoded earlier in the network, while higher level semantic information comes later. Furthermore, classifier performance generally increases for all tasks when more layers are accounted for, starting from layer 0, suggesting that useful information is being incorporated at each progressive layer. Others, such as \cite{van_Aken_2019}, looked specifically at QA with SQuAD and published similar findings. Their work suggests that different layers encode different information important for QA.

\cite{ma2019universal} showed that a simple averaging of only the first and last layers of BERT results in contextualized embeddings that performed better than only using the final layer. The authors evaluated this approach on a variety of tasks such as text classification and question-answering. Together, these works suggest that the hidden state activations within BERT may contain unique information that can be used to augment the final layer. That said, the exact way of doing this requires further exploration.

In this work, we leverage the findings by Tenney and Ma as inspiration and develop a novel solution that accomplished two fundemental goals: $1.$ Reduce expensive BERT fine-tuning; $2.$ More importantly, we were able to do so and maintain or exceed BERT-level performance in the process. We accomplished this by extracting the information-rich hidden states from each encoder layer and used the full embeddings as training data. We demonstrate that, for two question-answering tasks, even our simple architectures can match BERT performance at a fraction of the fine-tuning cost. Our best model saves on one full epoch of fine-tuning, and performs better than BERT, suggesting that our approach offers a novel and more efficient alternative to fine-tuning until convergence.

	\begin{figure*}[!h]
\floatbox[{\capbeside\thisfloatsetup{capbesideposition={right,top},capbesidewidth=8cm}}]{figure}[\FBwidth]
{\caption{Architecture of our best model in relation to BERT. \textbf{Left}: BERT and its span annotation inference process. \textbf{Right}: For our model, BERT embeddings are first transformed through our custom adapter layer. Next, the last two dimensions are flattened. Optionally, a skip connection is added between the sequence outputs of the final BERT layer and this flattened representation. This is present in the best model discovered at 3/10 of an epoch, but was not necessary for the best model discovered at 1 full epoch. This tensor is then projected down to (386, 2) with a densely connected layer and split on the last axis into two model heads.  These represent the logits of the start-span and end-span position.}\label{fig:test}}
{\includegraphics[width=7.5cm]{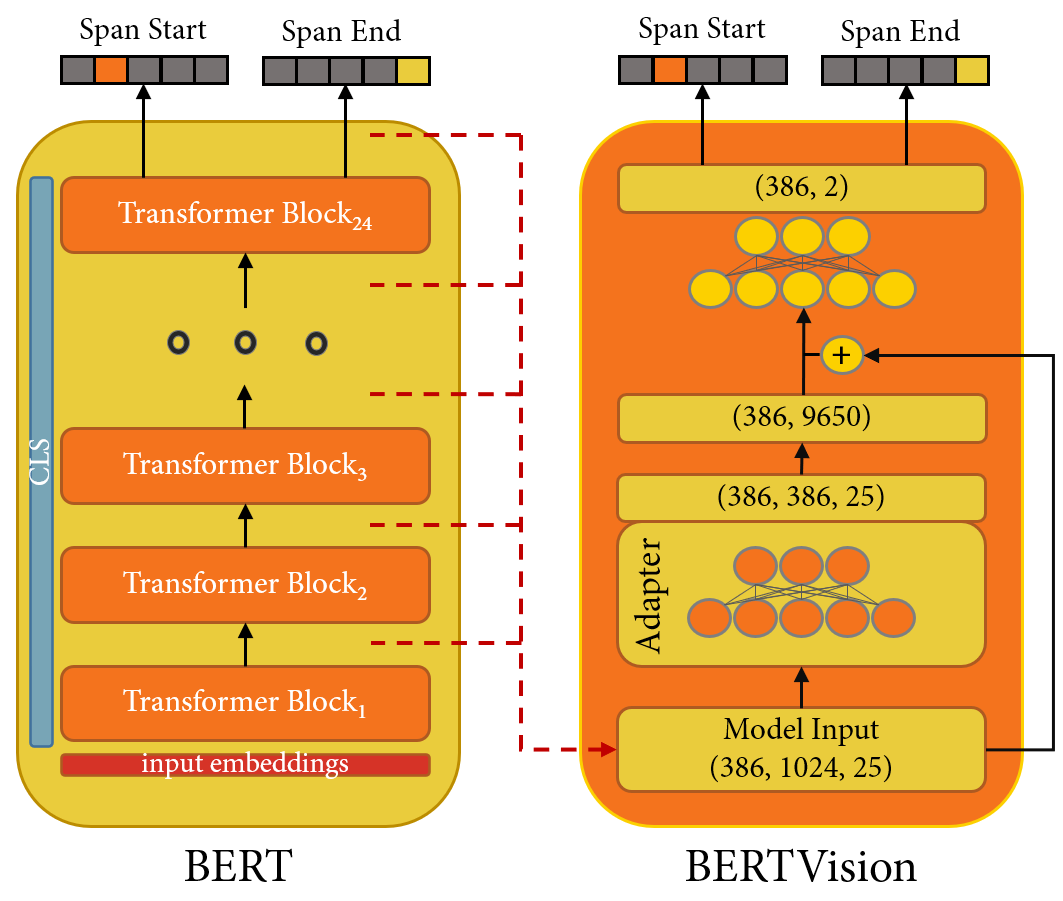}}
\end{figure*}
	
	\section{Methods}
\label{sec:methods}

This section introduces our baseline BERT model, and custom models trained on BERT embeddings. We also describe how we use the SQuAD 2.0 question-answering dataset for both span annotation and classification.

\subsection{Modeling approach}

Our custom models use the BERT activations from each encoder layer as input data. For a single SQuAD 2.0 example, our data point has a shape of (X,1024, 25), where $X=386$ for span annotation, and 1 for classification (see appendix [\ref{apdx:bertvision_span_annotation_data_pipeline_graph}] for details). We term this representation of the data as \textit{“embeddings.”}

\subsection{Learned and average pooling}

We implemented the pooling method described in \cite{tenney-etal-2019-bert}, which contracts the last dimension from 25 to 1 through a learned linear combination of each layer. We call this approach learned pooling (LP). We also evaluate the pooling approach reported in \cite{ma2019universal}, average pooling (AP), where each encoder layer is given equal weights. 

\subsection{Adapter compression}

We use the term “compression” to refer to methods that reduce the 1024 dimension of the BERT embeddings. The method proposed in \cite{DBLP:journals/corr/abs-1902-00751} is termed \textit{“adapter,”} which is an auto-encoder type architecture wherein the embeddings are first projected into a lower dimension using a fully-connected layer. For our purposes, the adapter serves as a bridge between BERT embeddings and downstream layers. The architecture in Houlsby et. al. can only handle 3D tensors (including batch), because each adapter handles data from a single transformer layer. Since we work with the activations from multiple layers at once, we need to be able to handle 4D tensors. To do this, our adapter implementation can treat each transformer layer as independent, learning separate compression weights for each layer. Alternatively, we can also employ weight-sharing, so that the compression for each encoder layer follows the same set of transformations.

\subsection{Custom CNNs}

We also implemented various novel CNN architectures, as well as modified existing models such as Inception and Xception \citep{DBLP:journals/corr/SzegedyLJSRAEVR14, DBLP:journals/corr/Chollet16a}. For span annotation, the key is that the CNN models must preserve the text length dimension of 386 (see appendix [\ref{apdx:cnn_models_preserving_seq_length}] for rationale). To view a notebook of all models we tried and their summaries, see: \href{https://github.com/cbenge509/BERTVision}{BERT Vision GitHub repo}.

\subsection{Fine-tuning BERT}

For all experiments, we use the $BERT_{LARGE}$ uncased implementation from \href{https://huggingface.co/}{Hugging Face}. To establish a baseline for our QA tasks, we fine-tuned BERT for 6 epochs with a setup similar to that described in \cite{Devlin2019}. For QA span annotation, questions that do not have answers have the start and end span positions assigned at the CLS token (position 0). We used the Adam optimizer with an initial learning rate of 1e-5. Due to hardware constraints, we use batch size of 4 rather than 48. At inference time, the most likely complete span is predicted based on the maximum softmax probability of the start- and end-span position. The setup is identical in classification, except that we used the pooled CLS token rather than the sequence outputs.
%\begin{equation} \label{eq1}
%Y_{\left(S,E\right)} = \left(\underset{start}{\arg\max} ,\; \underset{end}{\arg\max}\right)
%\end{equation}

\subsection{Ensembling} 

We evaluated multiple ensemble approaches for span annotation. The most successful method takes the element-wise max of the softmax probabilities output by each model in the ensemble. Let $Z$ and $Y$ be two model softmax probabilities for start or end span position. Then, the predicted position is:
\begin{equation} \label{eq2}
\begin{aligned}
\arg\max\left(\underset{{Z_i, Y_i}}{\max},\;\;\text{for i in \{1,2,...,386\}}\right)
\end{aligned}
\end{equation}

\subsection{Data processing and evaluation}

We use SQuAD 2.0 for our QA dataset. Standard Exact Match (EM) and F1-score (F1) are used for evaluation as outlined in \cite{DBLP:journals/corr/abs-1806-03822}. For classification, EM is equivalent to accuracy as we are predicting a single binary outcome. For our BERT models, we restrict the maximum token length to 386 (See appendix [\ref{apdx:explanation_max_sequence_length}] for rationale). Question-context pairs that exceed this maximum sequence are split into multiple segments (as many as needed) with an overlap between each segment of 128 tokens. For the splits that do not contain the answer, the labels for that split are set to “no answer.” At inference time, we take the argmax of the span probabilities predicted by each split as the final prediction for the example.
%\begin{equation} \label{eq3}
%EM = \begin{cases} \frac{1}{N}\sum_{i=1}^{N}{\left(\bar{s_i} = s_i,\;\bar{e_i} = e_i\right)} &\mbox{\small if QA} \\
%	\frac{1}{N}\sum_{i=1}^{N}{\left(\bar{s_i} = s_i\right)} &\mbox{\small else}
%	\end{cases} 
%\end{equation}
	\section{Results}

\subsection{Span Annotation}
For span annotation, we sought to generate useful embeddings with as little fine-tuning as possible. However, we were unable to find models that can fit BERT embeddings without some fine-tuning (see appendix [\ref{adpx:need_to_fine_tune}] for details). As a result, we decided that a small amount of fine-tuning would be necessary.

\subsubsection{BERT fine-tuning as baseline}
To establish a baseline, we fine-tuned BERT for up to 6 epochs with results shown in Figure [\ref{apdx:BERT_fine_tuning_span_annotation_sec}]. We measure performance for every 10th of a fractional epoch between 0 and 1 epochs, as well as full epochs up to 6. We observed that performance peaked at 2 epochs, achieving an Exact Match (EM) of 0.747, and an F1 score of 0.792. Between 0 and 1 epochs, performance consistently increased as measured by  both EM and F1. (For comparison with other published works, see appendix [\ref{apdx:comparison_with_devlin}]) For span detection we extracted embeddings at 3/10 of an epoch and at one full epoch. (for a rationale for this decision see appendix [\ref{apdx:explanation_of_input_embeddings_shape}])

\subsubsection{Models trained at 3/10 epoch embeddings}
Using the embeddings at 3/10 of an epoch, we explored over 20 parameter-efficient models using a combination of CNNs, pooling, and compression techniques. \footnote{See appendix [\ref{apdx:model_training_strategy}] for training time and strategy} Table [\ref{tbl:qa_3_10_Models}] shows that our best models outperform baseline BERT fine-tuned to 3/10 of an epoch. Results of all models are in appendix [\ref{apdx:span_annotation_all_results}].

First, we compare two models that use different pooling strategies, LP (learned pooling) and AP (average pooling) [\ref{sec:methods}]. We found that the AP model achieved similar performance as BERT itself, while LP improved performance. An analysis of the learned weights suggests that a non-uniform distribution of pooling weights is optimal, with the distribution favoring later layers of BERT compared to earlier layers (see appendix [\ref{apdx:non-uniform}]).

Next, we evaluated models leveraging adapters (see Methods). We found that our modified adapters of all flavors improved model performance compared to baseline BERT at 3/10 of an epoch, with our best model improving EM by 4.5 percentage points, and F1 by 3.8. As indicated in Table [\ref{tbl:qa_3_10_Models}], without weight-sharing, the number of parameters increases by $\sim$24x, with a penalty to model performance, making weight-sharing superior in every respect. In addition, \cite{DBLP:journals/corr/abs-1902-00751} found that an adapter size of 64 provides the best F1-score when used between transformer blocks. Our modified implementation did not perform well at this size with or without weight-sharing.

While we extensively explored stacking pooling with adapters and CNNs, we did not find a model which performed better than our best adapter model. Table [\ref{tbl:qa_3_10_Models}] shows the results, and the appendix [\ref{apdx:span_annotation_all_results}] contains the rest. For one model where we stacked a modified Xception network, the number of parameters increased by 16x, but performance was no better than pooling alone.
\begin{table}[ht]
	\centering
	\small
	\begin{tabular}{L{2.9cm}|C{1.5cm} C{0.9cm} C{0.9cm}}
		\toprule
		\textbf{Model} & \textbf{\% Params} & \multicolumn{2}{c}{\textbf{SQuAD2.0}}\\
		 & \textbf{BERT}$_{large}$ & \textbf{EM} & \textbf{F1}\\
		\midrule
		BERT $\frac{3}{10}e$ & 100\% & $0.654$ & $0.702$ \\
		learned pooling (LP) & 0.001\% & $0.675$ & $0.721$ \\
		average pooling (AP) & 0.001\% & $0.657$ & $0.700$ \\
		\textbf{adapter size 386} & \textbf{0.124}\% & \boldmath$0.699$ & \boldmath$0.740$ \\
		\hspace{0.5em} - weights not shared & 2.957\% & $0.676$ & $0.723$ \\
		adapter size 64 & 0.021\% & $0.680$ & $0.732$ \\
		\hspace{0.5em} - weights not shared & 0.491\% & $0.684$ & $0.716$ \\
		LP adapter size 386 \& xception & 1.970\% & $0.668$ & $0.711$ \\
		\bottomrule
	\end{tabular}
	\caption{\label{tbl:qa_3_10_Models}Models trained on embeddings at $\frac{3}{10}$ epochs}
\end{table}

\subsubsection{Best models on top at 1 epoch}
Using 1 epoch embeddings as our training data, we fit the same set of models as with the 3/10 epoch embeddings (see the appendix [\ref{apdx:span_annotation_all_results}] for full results). In most cases, we found that our models outperformed BERT at the same level of fine-tuning. The best model is nearly identical to that found with 3/10 epoch embeddings (see table [\ref{tbl:qa_dev_set_performance}]), outperforming BERT by 2.1 percentage points in EM, and 1.3 in F1. This model’s performance is also competitive with maximum performance achieved with BERT on this task, which is BERT fine-tuned to 2 epochs. This result shows that we are able to reduce BERT fine-tuning by 1 epoch and still achieve comparable performance with a model of only about 0.124\% of the number of BERT parameters. The implications of our findings suggest that near-optimal BERT performance can be achieved in a fraction of the training time and GPU/TPU expense. 
\begin{table}[ht]
	\centering
	\small
	\begin{tabular}{L{2.9cm}|C{1.5cm} C{0.9cm} C{0.9cm}}
		\toprule
		\textbf{Model} & \textbf{\% Params} & \multicolumn{2}{c}{\textbf{SQuAD2.0}}\\
		& \textbf{BERT}$_{large}$ & \textbf{EM} & \textbf{F1}\\
		\midrule
		BERT $1e$ & 100\% & $0.728$ & $0.777$ \\
		BERT $2e$ (\textit{best}) & 100\% & $0.747$ & $0.792$ \\
		\textbf{adapter size 386} & \textbf{0.124}\% & \boldmath$0.749$ & \boldmath$0.790$ \\
		\hspace{0.5em} - weights not shared & 2.957\% & $0.716$ & $0.767$ \\
		adapter size 64 & 0.021\% & $0.739$ & $0.785$ \\
		\hspace{0.5em} - weights not shared & 0.491\% & $0.736$ & $0.784$ \\
		LP adapter size 386 \& xception & 1.970\% & $0.741$ & $0.786$ \\
		\bottomrule
	\end{tabular}
	\caption{\label{tbl:qa_dev_set_performance}Models trained on embeddings at $1$ epoch}
\end{table}

\subsubsection{Training with less data}
\begin{figure}[ht]
	\centering
	\includegraphics[width=7.5cm]{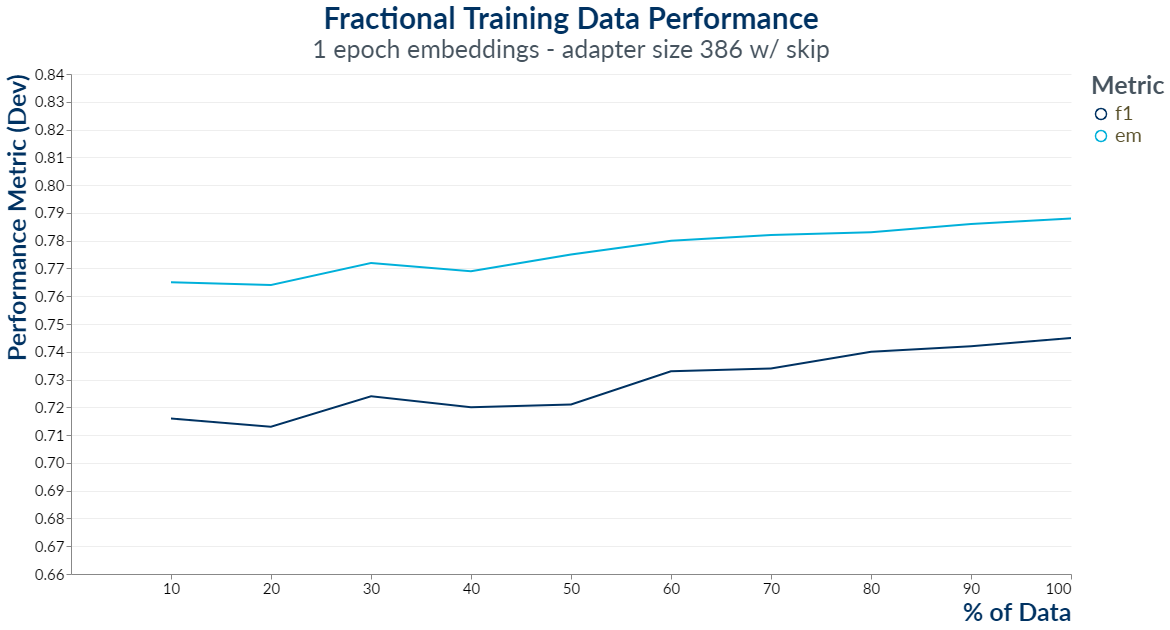}
	\caption{\label{fig:1_epoch_embeddings__adapter_386_with_skip}Training with less data}
\end{figure}

The previous sections show that our models perform well compared to BERT when trained on the full dataset. However, since supervised data can be difficult to obtain, we explore whether the same performance can be achieved with significantly less data. To answer this question, we trained our best models on varying amounts of data, ranging from 0\% to 100\%. As with the full dataset, all models were trained for 1 epoch. Figure [\ref{fig:1_epoch_embeddings__adapter_386_with_skip}] shows the results. In both cases, we rapidly approach strong performance early, beating BERT at $\sim$30\% of the data. Even by 10\%, we approach peak performance. This shows that we can achieve strong performance even when using less training data. 

\subsubsection{Ensembling our models with BERT}
Since our approach requires fine-tuning BERT in order to derive the embeddings we use as the input layer in our models, we get BERT predictions "for free."  In other words, we have access to the full BERT model for predictions as well as our own embeddings-based models. Thus, we can ensemble our models with BERT predictions with no additional training required. Table [\ref{tbl:qa_ensembling}] presents the results of this exercise. Ensembling the models at 3/10 of an epoch did not lead to an improvement, but ensembling our 1 epoch model with BERT at 1 epoch led to more than a half percentage point improvement. This suggests that our model and BERT supply complementary information that together leads to better predictions.

\begin{table}[ht]
	\centering
	\small
	\begin{tabular}{L{4.2cm}|C{0.9cm} C{0.9cm}}
		\toprule
		\textbf{Model} & \multicolumn{2}{c}{\textbf{SQuAD2.0}}\\
		& \textbf{EM} & \textbf{F1}\\
		\midrule
		BERT $\frac{3}{10}e$     					& $0.654$  & $0.702$ \\
		our model $\frac{3}{10}e$ 					& $0.699$  & $0.740$ \\
		BERT $1e$     								& $0.728$  & $0.777$ \\
		our model $1e$ 								& $0.749$  & $0.790$ \\
		ensemble BERT+our model $\frac{3}{10}e$		& $0.691$  & $0.734$ \\
		\textbf{ensemble BERT+our model} \boldmath$1e$  & \boldmath$0.756$  & \boldmath$0.798$ \\
		\bottomrule
	\end{tabular}
	\caption{\label{tbl:qa_ensembling}QA ensembling results}
\end{table}

\subsection{Classification}

The previous section on QA span annotation uses the full sequence embeddings generated by BERT. Another common way to apply BERT is to text classification, which only uses the CLS token. Our goal in this section is to explore the efficacy of leveraging the CLS token hidden state activations in an analogous manner to our approach with the span annotation task. 

\subsubsection{BERT fine-tuning as baseline}

Similar to QA span annotation, we use BERT fine-tuning as a baseline. We fine-tuned BERT for 6 epochs using the CLS token, with the EM and F1 for each epoch shown in Figure [\ref{apdx:BERT_fine_tuning_classification}]. Our best performance was achieved at 3 epochs. We utilized embeddings at 2/10 of an epoch and 1 full epoch for the binary classification task (see appendix [\ref{adpx:need_to_fine_tune}] for rationale).

\subsubsection{Models trained at 2/10 epoch embeddings}

\begin{figure}[ht]
	\centering
	\includegraphics[width=7.5cm]{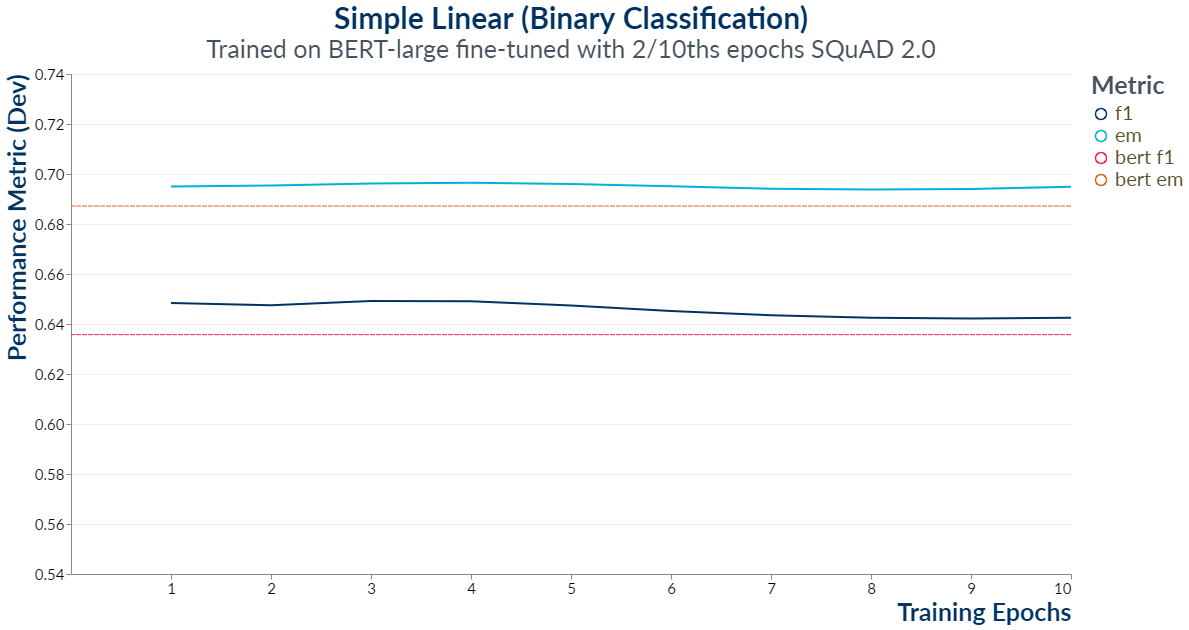}
	\caption{\label{fig:bc_2_10ths_performance}2/10ths epoch model vs BERT}
\end{figure}

At 2/10ths of an epoch, BERT achieved an F1 of 0.635 and EM of 0.687 [\ref{apdx:BERT_fine_tuning_classification}]. We tried training various parameter-efficient model architectures similar to span annotation, most of which were CNN-based or simple linear. Our best performing model leverages a simple linear weighting that contracts across the channels dimension through learned weights inspired by \cite{tenney-etal-2019-bert}. We trained for 10 epochs and recorded performance of the model at each epoch evaluated against the dev set. At every epoch, the model outperformed the BERT baseline at 2/10 of an epoch [\ref{fig:bc_2_10ths_performance}].

\subsubsection{Models trained at 1 epoch embeddings}

With 1 epoch embeddings, our results show significant improvement over the 2/10ths epoch models. Surprisingly, the best results were achieved at 1 epoch of training with the simple linear model consisting of only 0.001\% of BERT’s parameters. This model outperforms BERT at both 1 and 2 epochs of fine-tuning. This implies that our parameter efficient model can save on 1 full epoch of BERT fine-tuning. However, performance does fall short of maximal BERT performance achieved at 3 epochs, and we were unable to find a model architecture that achieves this [\ref{tbl:bc_best_models}]. Nevertheless, this suggests that with text classification, we can use the CLS token hidden state activations in the same way we used the full sequence hidden state activations in the span annotation task. 

\begin{table}[ht]
	\centering
	\scriptsize
	\begin{tabular}{C{0.17cm}|C{0.7cm} C{0.7cm}|C{0.7cm} C{0.7cm}|C{1cm} C{1cm}}
		\toprule
		\boldmath$e$ & \textbf{BERT F1} & \textbf{BERT EM} & \textbf{Our F1} & \textbf{Our EM} & \textbf{F1 $\Delta$} & \textbf{EM $\Delta$} \\
		\midrule
		$1$ & \textcolor{berkeleyblue}{$0.720$} & \textcolor{berkeleyblue}{$0.761$} & $0.763$ & $0.782$ & \textcolor{laplane}{\boldmath$+0.043$} & \textcolor{laplane}{\boldmath$+0.021$} \\
		$3$ & \textcolor{berkeleyblue}{$0.790$} & \textcolor{berkeleyblue}{$0.804$} & $0.795$ & $0.811$ & \textcolor{laplane}{\boldmath$+0.005$} & \textcolor{laplane}{\boldmath$+0.007$} \\
		\bottomrule
	\end{tabular}
	\caption{\label{tbl:bc_best_models}Comparison of BERT and our models performance at 1 and 3 epochs on binary classification task}
\end{table} 

\subsubsection{Modeling with 3 full epoch embeddings}

BERT achieved its best performance against the SQuAD 2.0 binary classification task at 3 full epochs (0.79 F1, 0.804 EM) [\ref{apdx:BERT_fine_tuning_classification}].  Given that the simpler linear model outperformed BERT at the same level of fine-tuning, we wanted to evaluate if the simple model’s trend in performance would continue as BERT exhausts its ability to learn the binary classification task.  In this final evaluation, the linear model again outperformed BERT, achieving the top overall score for our task, albeit at a smaller margin [\ref{tbl:bc_best_models}].
	\section{Model analysis}

\subsection{Question types}

\begin{figure}[ht]
	\centering
	\includegraphics[width=7.5cm]{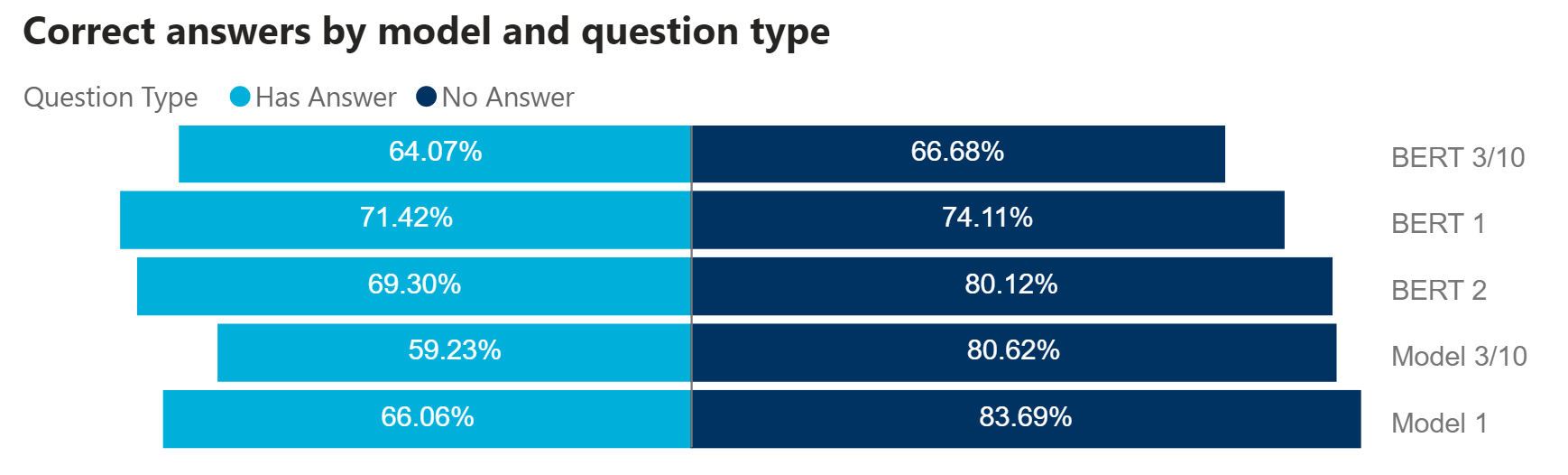}
	\caption{\label{fig:qa_correct_answers_by_model_and_type}Correctly answered percentages by model}
\end{figure}

Model performance on SQuAD 2.0 can be categorized based on whether the question-context pair has an answer (Has Answer versus No Answer). The dev set is very balanced in this sense as it contains 5,928 ($\sim$49.93\%) pairs with answers and 5,945 ($\sim$50.07\%) pairs without answers. Figure [\ref{fig:qa_correct_answers_by_model_and_type}] shows the fraction of questions of each type that BERT and our best-performing models correctly answers. As BERT is fine-tuned, the number of correctly answered questions in both categories increases, but questions with “No Answer” increases more rapidly. Our best models answer more “No Answer” questions correctly than BERT, but underperform BERT in terms of “Has Answer” questions.

To investigate further, we looked at our 3/10 epoch model’s incorrect predictions on “Has Answer” questions, and found that a large percentage, $\sim$65\%, were predicted as having no answer (rather than with the wrong answer). These results suggest that our models may be more liberal than BERT at predicting “no answer.”

\subsection{Second most likely answer}

For questions where our 3/10 epoch model’s most likely answer was incorrect, we considered the second most likely answer predicted by the model. We found that the second most likely answer is correct 44\% of the time when the most likely answer is incorrect. We also note 47\% of these were also “Has Answer” questions, which is a more balanced distribution compared to the most likely answer (where only 42\% of correctly answered questions have an answer). This suggests that even though our model at 3/10 is liberal at predicting no answer, its second most likely answer is quite frequently correct and less biased. 

\subsection{Context Length}

The previous section shows that our models are better at recognizing questions with no answers compared to BERT. Here, we look at which models perform better on longer sequences. At 3/10 of an epoch of fine-tuning, BERT and our models performed similarly (BERT average length: 165.86, our model average length: 164.50). However, we believe this is because both models answered many of the same questions correctly. When looking at questions that were uniquely answered correctly by each model, a clear difference emerges: the average sequence length for BERT is longer (BERT: 175.30, our model: 161.58). This suggests that our models are better at answering questions with shorter contexts than BERT. The same trend holds at 1 epoch of fine-tuning.

\subsection{Do 2 epochs perform even better?}
	
Our previous results show that models built on BERT embeddings at 1 epoch achieves maximal BERT performance at 2 epochs. Here, we investigate if we can further improve performance by training directly on embeddings derived at 2 epochs. While this approach does not reduce BERT fine-tuning time, it does provide a way to investigate whether our approach can achieve even better performance with longer fine-tuning. 
	
Table [\ref{tbl:bc_bert_fine_tuning}] shows the results. As expected, the 2-epoch models outperform the same models trained on 1-epoch embeddings. Since ensembling improved performance at 1 epoch, we applied the same approach to the models at 2 epochs. Surprisingly, the 2 epoch ensemble does not outperform the 1 epoch ensemble in terms of either EM (0.749 vs 0.756) or F1 (0.795 vs 0.798). This suggests that additional fine-tuning does not guarantee better performance. Our 1-epoch models, in addition with ensembling, not only saves on BERT fine-tuning time, but also achieves the best possible performance, suggesting a total lack of need for fine-tuning beyond 1 epoch.

\begin{table}[ht]
	\centering
	\small
	\begin{tabular}{L{4.2cm}|C{0.9cm} C{0.9cm}}
		\toprule
		\textbf{Model} & \multicolumn{2}{c}{\textbf{SQuAD2.0}}\\
		& \textbf{EM} & \textbf{F1}\\
		\midrule
		BERT $2e$ (\textit{best}) 		& $0.747$ & $0.792$ \\
		best model architecture from $1e$	& $0.753$ & $0.797$ \\
		best model architecture from $\frac{3}{10}e$ 		& $0.749$ & $0.795$ \\
		ensemble BERT $2e$ + best $1e$	& $0.751$ & $0.796$ \\
		\bottomrule
	\end{tabular}
	\caption{\label{tbl:bc_bert_fine_tuning}Models trained on embeddings at 2 epochs}
\end{table}

	\section{Conclusion and future work}

In this paper, we propose a parameter-efficient approach that achieves maximal BERT performance for QA span annotation and greatly reduces fine-tuning time required for BERT. Our models are trained on BERT hidden state activations (embeddings), and consistently outperform BERT at the same level of fine-tuning. By using an ensemble of our model with BERT’s predictions, we further surpass BERT performance, reducing the need for fine-tuning by 1 epoch. We achieved similarly promising results for QA classification, which suggests that this approach works well with both the full sequence BERT embeddings and with the CLS token embedding. Future work might look at reducing fine-tuning even further, by focusing on modeling, or alternative approaches for combining BERT embeddings. Better data caching strategies could also help the practical application of this method in production, as data loading from disk can be slow for generic hardware. While this work focused on SQuAD 2.0, further work can also evaluate this method on other NLP tasks, such as GLUE. Success here would demonstrate the ability of the method to generalize beyond QA, including for advanced inference tasks, such as entailment and paraphrasing. Further, to evaluate our relative success, we must evaluate our methods against current state-of-the-art approaches to reducing BERT size, while attempting to retain performance. Overall, we demonstrated that BERT embeddings carries valuable information that can be leveraged for inference, in a way that reduces BERT fine-tuning and exceed BERT performance.

	% ---------------------------------------------------------------------------
	% formatting examples; we will throw this out when paper is done.
	% ---------------------------------------------------------------------------
	%\input{formattingexamples}
	% ---------------------------------------------------------------------------

	\section*{Acknowledgments}

Special thanks to Daniel Cer and Joachim Rahmfeld for their feedback and guidance.

	%\nocite{*}
	\bibliography{references}
	\bibliographystyle{acl_natbib}

	\appendix

\section{Appendices}
\label{sec:appendix}
\small

\subsection{BERT fine-tuning figures}
\label{apdx:BERT_fine_tuning_span_annotation_sec}

BERT$_{large}$ was fine-tuned for both span annotation and classification tasks; below are tables and blots indicating the performance for each task.

\begin{figure}[h]
	\centering
	\begin{subfigure}{0.95\textwidth}%
		\centering
		\includegraphics[width=\linewidth]{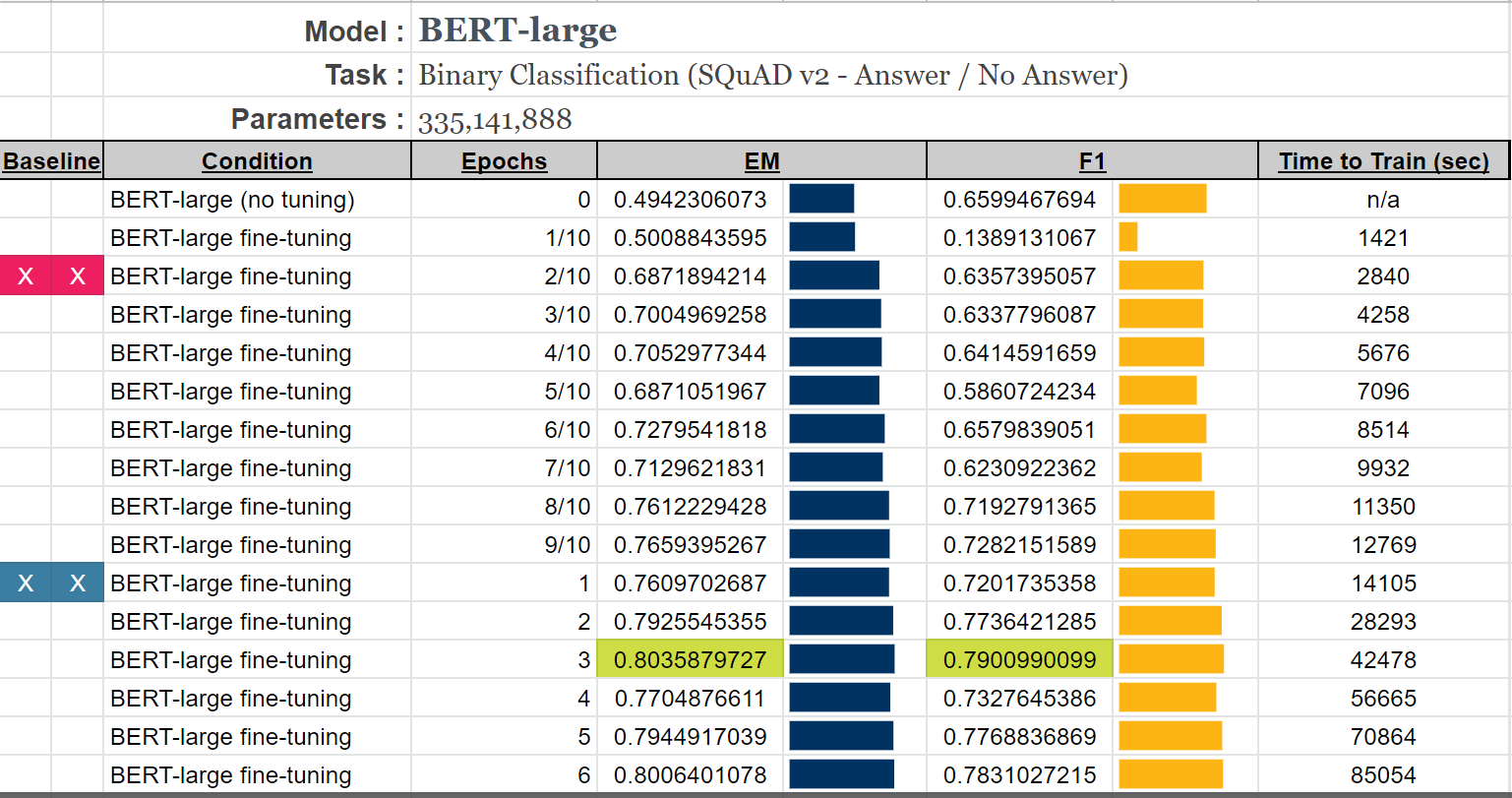}%
		\caption{BERT$_{large}$ fine-tuning table : classification}
	\end{subfigure}%

	\vspace*{8pt}%
	
	\begin{subfigure}{0.96\textwidth}%
		\centering
		\includegraphics[width=\linewidth]{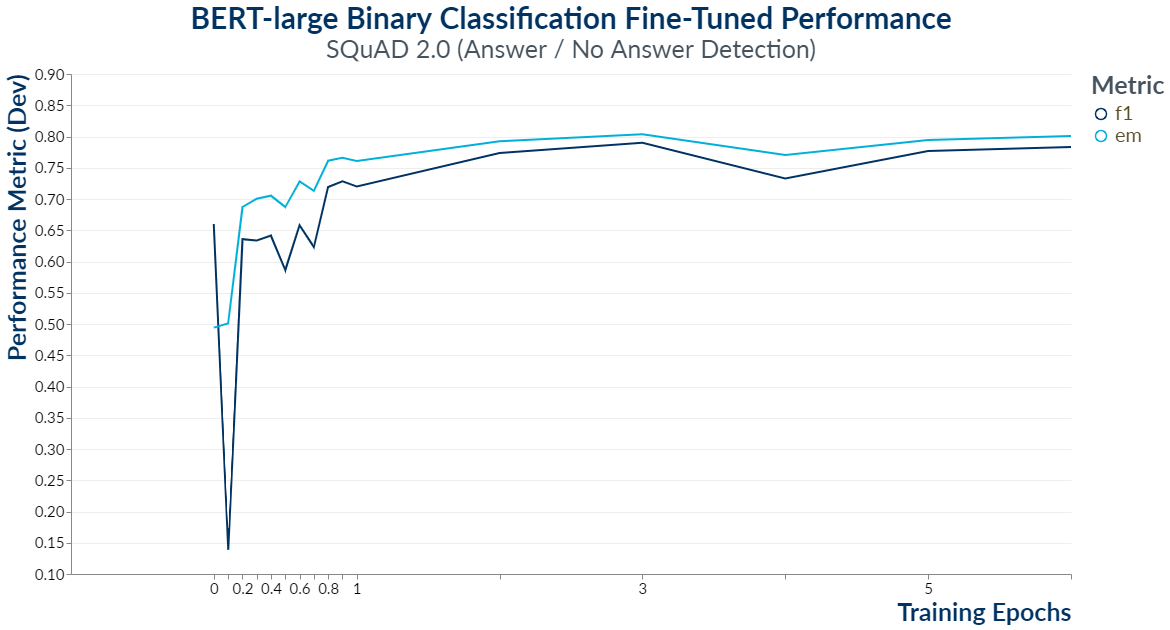}%
		\caption{BERT$_{large}$ fine-tuning plot : classification}
	\end{subfigure}%
	\caption{\label{apdx:BERT_fine_tuning_classification}BERT$_{large}$ fine-tuning for classification}
\end{figure}%

\begin{figure}[!h]
	\centering
	\begin{subfigure}{0.95\textwidth}%
		\centering
		\includegraphics[width=\linewidth]{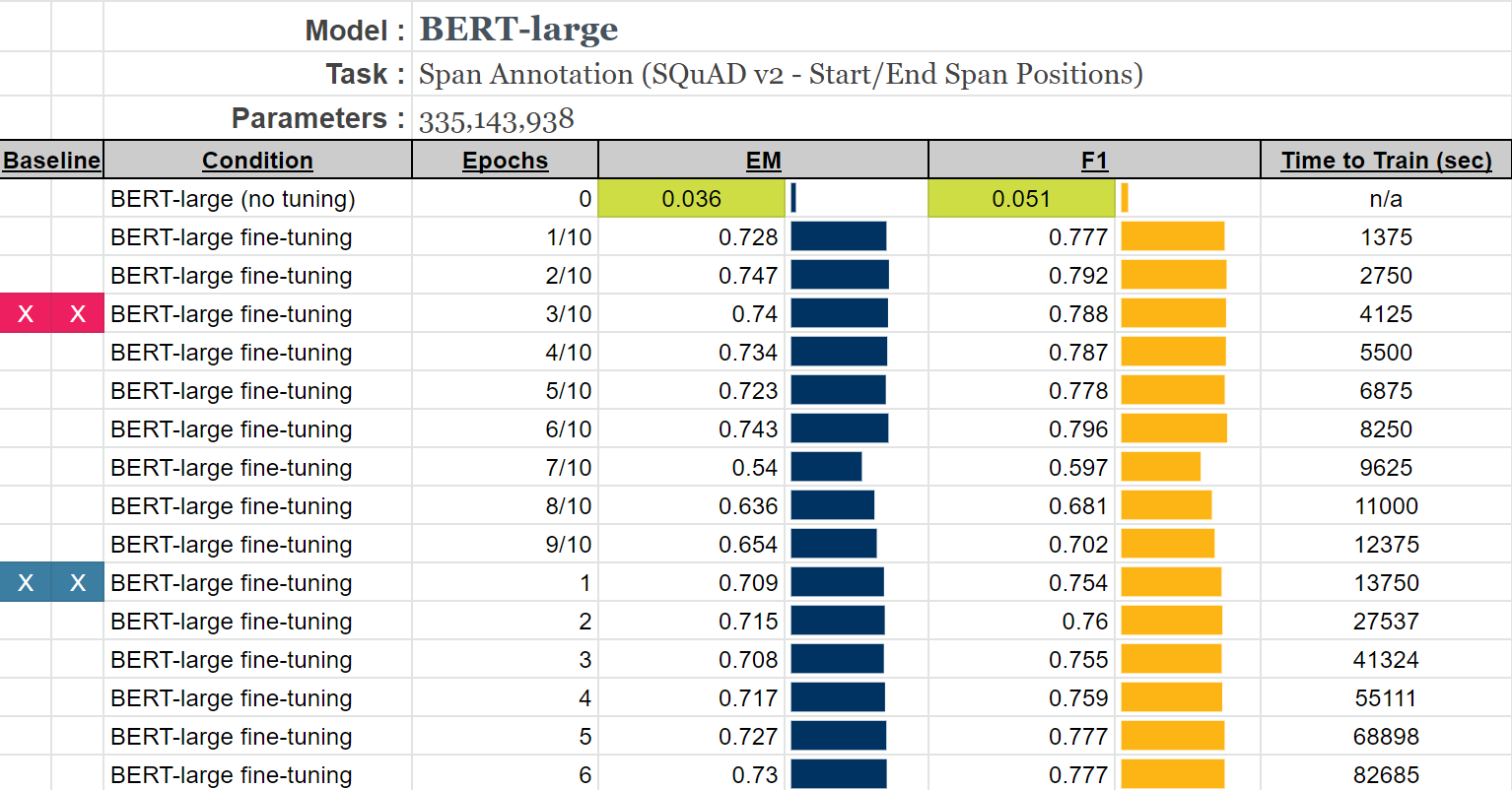}%
		\caption{BERT$_{large}$ fine-tuning table : span annotation}
	\end{subfigure}%
	
	\vspace*{8pt}%

	\begin{subfigure}{0.96\textwidth}%
		\centering
		\includegraphics[width=\linewidth]{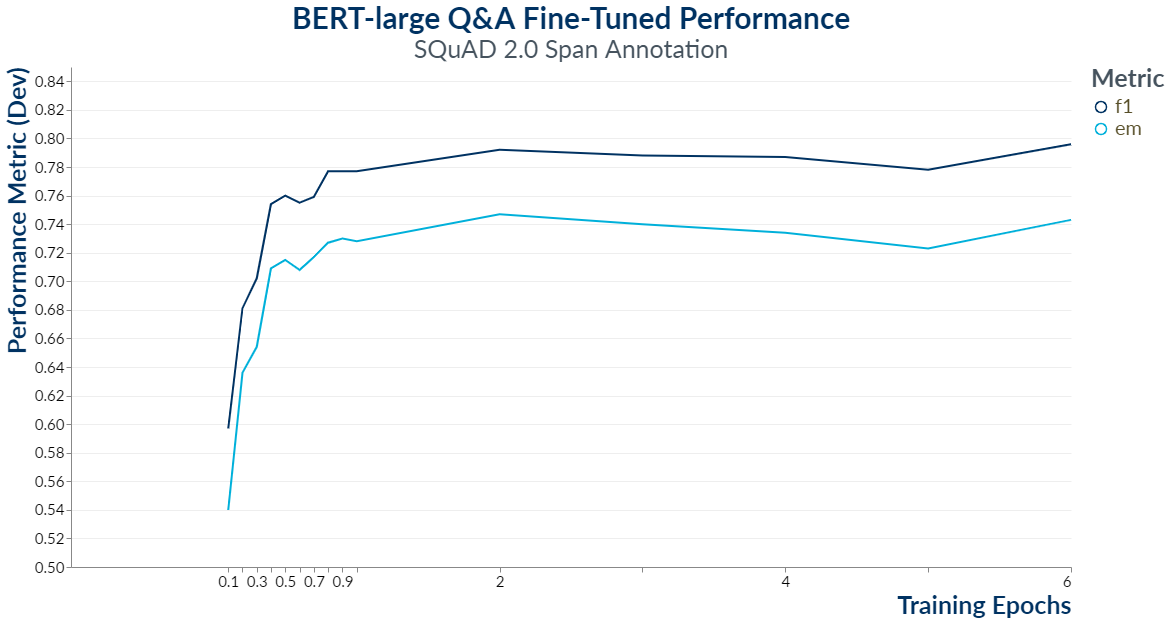}%
		\caption{BERT$_{large}$ fine-tuning plot : span annotation}
	\end{subfigure}%
	\caption{\label{apdx:BERT_fine_tuning_span_annotation}BERT$_{large}$ fine-tuning for span annotation}
\end{figure}

\newpage
\subsection{Table of all model results for span annotation}
\label{apdx:span_annotation_all_results}

Models were trained on embeddings derived from 3/10 of an epoch and 1 full epoch.  Performance on evaluation of the dev set is presented below.

\begin{figure}[h]
	\centering
	\includegraphics[width=\linewidth]{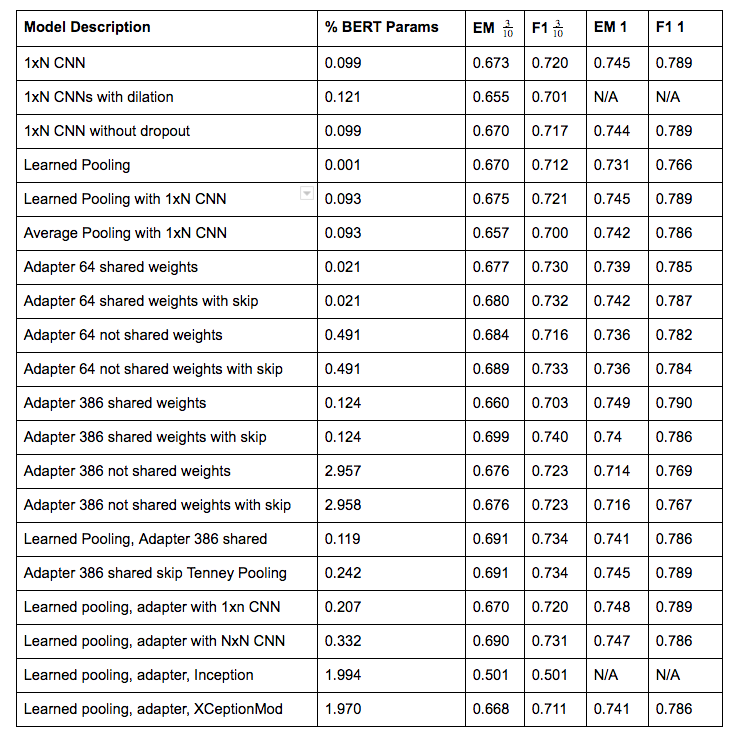}%
	\caption{BERT$_{large}$ All model performance : span annotation}
\end{figure}

\subsection{Best model performance for classification task}

The best model evaluated for the classification task was a simple linear model that performed channel contraction using learned weights followed by a simple dense layer connection with no softmax.  As indicated below, this model outperformed BERT$_{large}$ for almost every metric save for F1 on 3 epoch fine-tuned embeddings.

\begin{figure}[h]
	\centering
	\includegraphics[width=\linewidth]{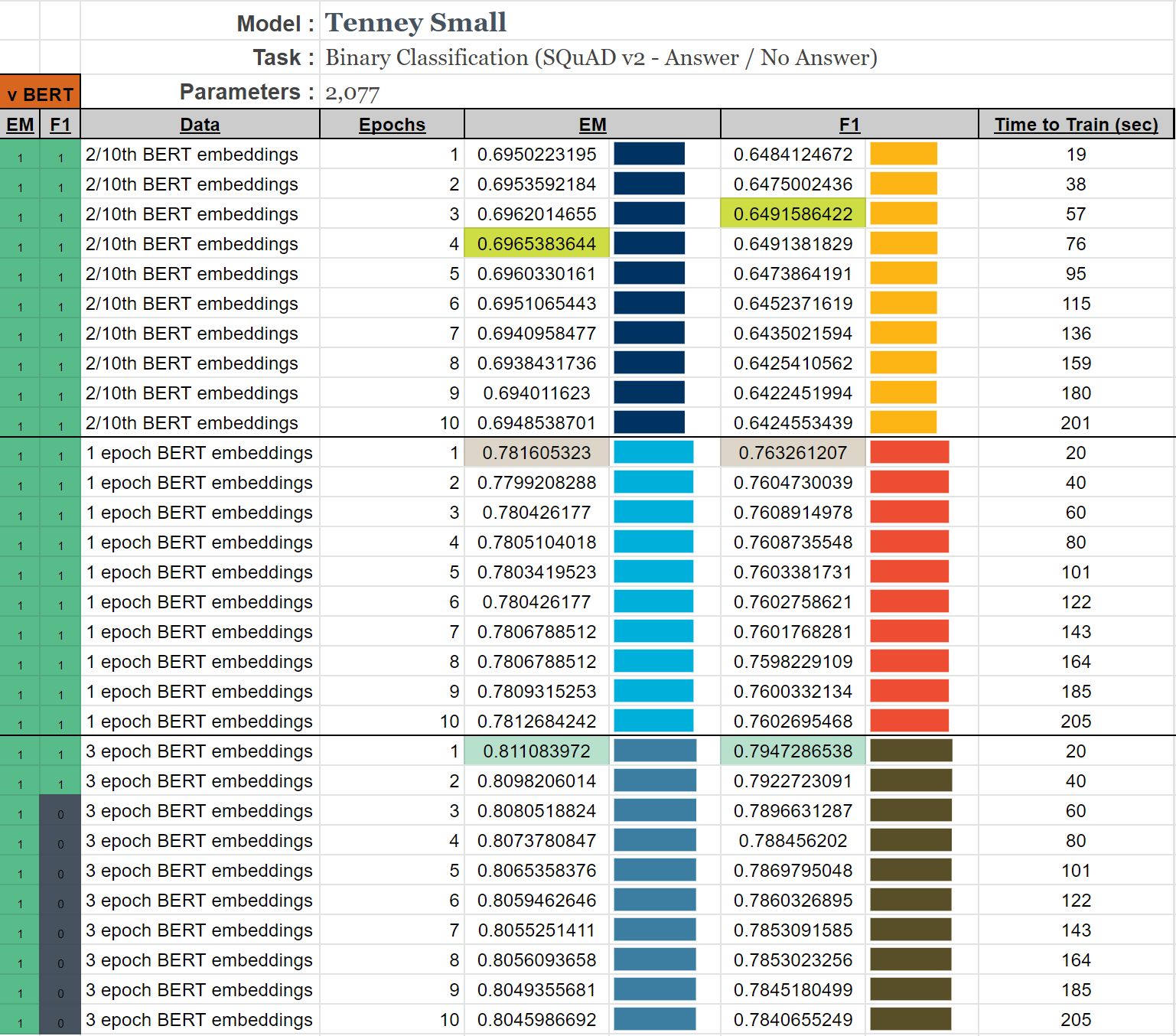}%
	\caption{BERT$_{large}$ Best model performance : classification task}
\end{figure}

\newpage

\begin{figure*}[t]
	\centering
	\includegraphics[width=\textwidth]{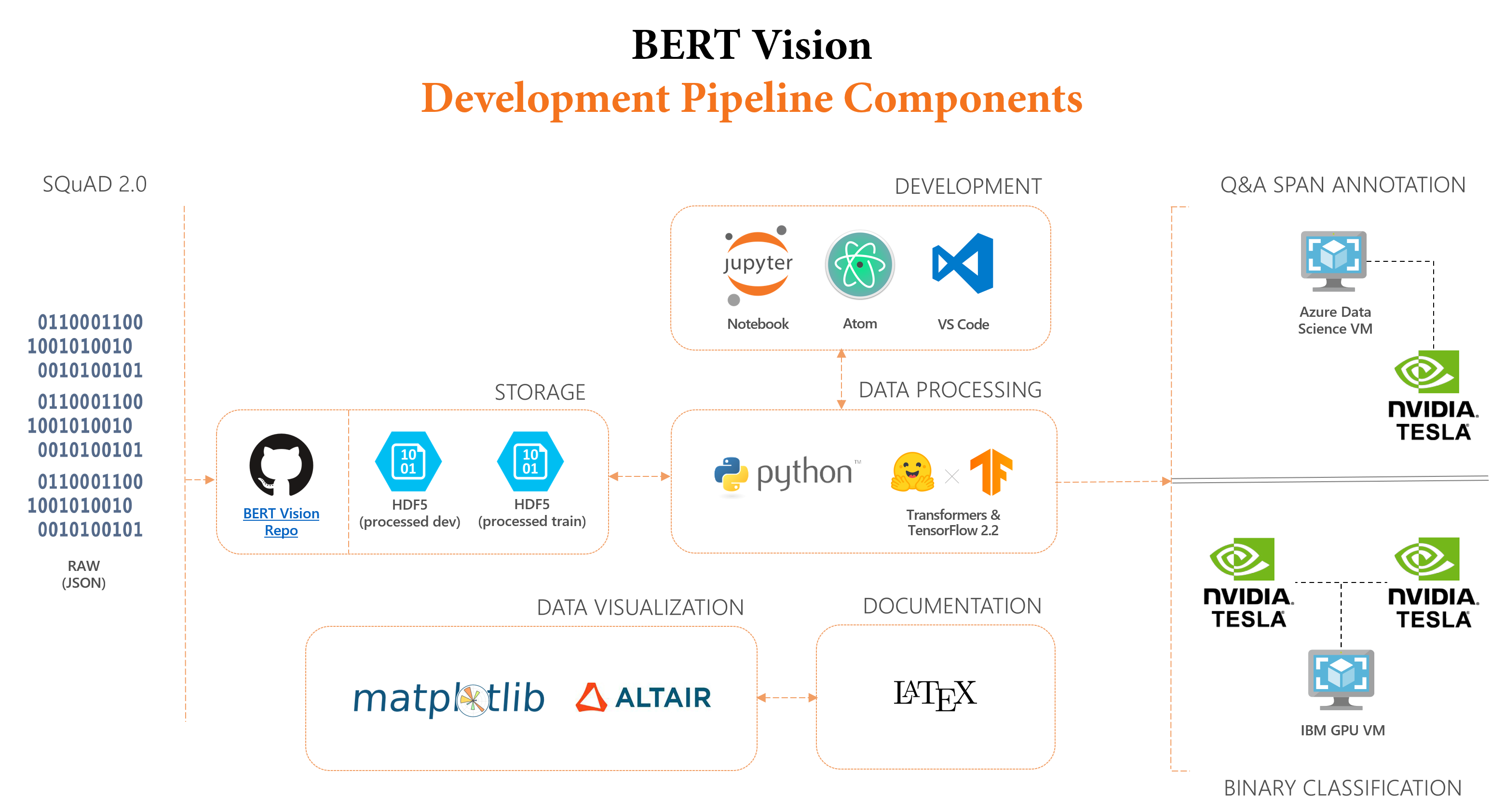}%
	\caption{BERTVision development pipeline}
	\label{apdx:bertvision_development_pipeline_graph}
\end{figure*}

\newpage

\subsection{Table of all model results for classification}
\label{apdx:classification_models_trained}

Models were trained on BERT$_{large}$ fine-tuned embeddings derived from 2/10 of an epoch and 1 full epoch. Additionally, the most successful model was trained on 3 epochs fine-tuned embeddings. Performance on evaluation of the dev set is presented below.

\begin{table}[h]
	\centering
	\small
	\begin{tabular}{L{2.9cm}|C{1.5cm} C{0.9cm} C{0.9cm}}
		\toprule
		\textbf{Model} & \textbf{\% Params} & \multicolumn{2}{c}{\textbf{SQuAD2.0}}\\
		& \textbf{BERT}$_{large}$ & \textbf{EM} & \textbf{F1}\\
		\midrule
		adapter pooler tenney 	& 0.119\% & 0.685 & 0.646 \\
		xception abbr 			& 0.079\% & 0.673 & 0.702 \\
		\textbf{xception}		& \textbf{6.181\%} & \textbf{0.707} & \textbf{0.710} \\
		xception abbr cls 		& 0.080\% & 0.672 & 0.564 \\
		adapter pooler avg 		& 0.119\% & 0.691 & 0.624 \\
		tenney small 			& 0.001\% & 0.697 & 0.650 \\
		\bottomrule
	\end{tabular}
	\caption{Models trained on embeddings at $\frac{2}{10}e$}
\end{table}

\begin{table}[h]
	\centering
	\small
	\begin{tabular}{L{2.9cm}|C{1.5cm} C{0.9cm} C{0.9cm}}
		\toprule
		\textbf{Model} & \textbf{\% Params} & \multicolumn{2}{c}{\textbf{SQuAD2.0}}\\
		& \textbf{BERT}$_{large}$ & \textbf{EM} & \textbf{F1}\\
		\midrule
		adapter pooler tenney 	& 0.119\% & 0.781 & 0.765 \\
		\textbf{xception abbr}	& \textbf{0.079\%} & \textbf{0.785} & \textbf{0.782} \\
		xception 				& 6.181\% & 0.783 & 0.766 \\
		xception abbr cls 		& 0.080\% & 0.783 & 0.780 \\
		adapter pooler avg 		& 0.119\% & 0.779 & 0.765 \\
		tenney small 			& 0.001\% & 0.782 & 0.763 \\
		\bottomrule
	\end{tabular}
	\caption{Models trained on embeddings at $1e$}
\end{table}

\begin{table}[h]
	\centering
	\small
	\begin{tabular}{L{2.9cm}|C{1.5cm} C{0.9cm} C{0.9cm}}
		\toprule
		\textbf{Model} & \textbf{\% Params} & \multicolumn{2}{c}{\textbf{SQuAD2.0}}\\
		& \textbf{BERT}$_{large}$ & \textbf{EM} & \textbf{F1}\\
		\midrule
		tenney small & 0.001\% & 0.811 & 0.795 \\
		\bottomrule
	\end{tabular}
	\caption{Models trained on embeddings at $3e$}
\end{table}

\subsection{Data challenges and training time}
\label{apdx:data_challenges}

During training, in order to obtain the internal BERT embeddings for each example (which is the input “data” into our downstream models), we had to either: 1.  Pre-generate the embeddings, or 2. Generate the embeddings for each example on the fly from a frozen BERT model. Due to the size of BERT, we quickly ran out of GPU memory with the second method, so we had to resort to the first. Working with BERT embeddings on the SQuAD 2.0 data set presented a data engineering problem due to the size of the data. Using the 4-byte float representation, the entire dataset with each example having shape (386, 1024, 25) is approximately 5 TB in size, which is far too large to store into memory on our hardware. Instead, this data was written to disk, and a custom Keras data generator used to retrieve this data in batch sizes of 16 for training (in a shuffled random order).

While this method presents no cost to accuracy, the I/O time for loading the data was significant, much longer in most cases than the actual training time for fitting any of our models. For all models, training per epoch was around 5 hours, but we estimate that on average over 95\% of the time is spent on loading data, and only 5\% of the time fitting the model. As a result, in this work, in order to separate out infrastructure issues with model training, as a proxy for training cost, we compare the number of parameters in our model rather than wall clock training time. We also explore the potential to use less embedding data during fitting, and for future work, a faster storage system should be explored in order to advance this work for practical applications. 

We also note that for the classification task, we did not experience any data management issues. Since we only use the CLS token for classification rather than the entire sequence length of 386, the full embeddings of all examples were a little over 13GB, which easily fit into CPU memory. Training for each epoch for on the order of minutes rather than hours.

\subsection{Model training strategy}
\label{apdx:model_training_strategy}

We train all of these models for a single epoch for three reasons: 1. Performance was already desirable at a single epoch. 2. Further training typically did not help performance further, 3. Due to our data management issue, loading the data usually takes more than 95\% of the training time, which makes it difficult to train for extended numbers of epochs (See [\ref{apdx:data_challenges}]). 

\subsection{Explanation on the shape of input embeddings data}
\label{apdx:explanation_of_input_embeddings_shape}

For span annotation, for a single SQuAD 2.0 example, the data point has a shape of (386,1024,25). Here, 386 represents the text length dimension, and 1024 the BERT embedding dimension for each encoder layer. The 25 comes from the stacking of the contextual wordpiece \cite{DBLP:journals/corr/WuSCLNMKCGMKSJL16} text embeddings, the 23 hidden state activations, and the final sequence outputs (24th layer) for BERT. 

For classification, an example has input shape (1,1024, 26). Here, 1 represents the lone CLS token, 1024 the BERT embedding dimension. The 26 comes from the stacking of the contextual wordpiece text embeddings, 23 hidden state activations, the final CLS token for the output encoder layer (24th layer), and the pooled CLS token from the final pooler layer.

\subsection{Explanation on max sequence length input into BERT}
\label{apdx:explanation_max_sequence_length}

BERT$_{large}$ has a maximum input sequence length of 512. While we could have truncated our question-context pairs at this length, we found that a large majority of examples were much shorter than this length. The average length of the input is about 171 tokens long. The length we chose was 386, which is between 98-99th percentile. As a result, without much loss, we can significantly save on the number of mostly meaningless “[PAD]” tokens we needed to store, especially for full 25-layers of BERT embeddings.

\subsection{CNN models preserving max sequence length}
\label{apdx:cnn_models_preserving_seq_length}

Our CNN models were all designed to preserve the sequence length of the input question-context pair. This was achieved with 1xN convolutions so that these filters only compress along the 1024 dimension, or with NxN convolutions with padding along the sequence token dimension. In the case of 1xN convolutions, this is similar to a unigram model that treats each token separately. The NxN convolutions were n-gram models, ranging from 1 - 7 (depending on the size of N). The reason we did this is because we also tried convolution blocks from Inception [1] that gradually shrinks the 386 dimension; however, this model failed to learn to even fit the training data. As a result, we believe that for span annotation, since the data starts with a 386 dimension representing token position, and ends by predicting a probability for each position, we need this sentence length dimension throughout the entire model. A traditional computer vision CNN model such as Inception shrinks the image along this dimension, which destroys the structure necessary for span detection. Shrinking the text length and later expanding it again loses information, resulting in a failure to fit the data.

\subsection{Comparison of our BERT performance with Devlin}
\label{apdx:comparison_with_devlin}

For our fine-tuning procedure, we observed that performance peaked at 2 epochs, achieving an Exact Match (EM) of 0.747, and an F1 score of 0.792. This is slightly worse than that reported by \cite{Devlin2019} at an EM of 78.7 and F1 of 81.9. We hypothesize that these difference might arise due to our training batch size, random initializations, and the fact we do not favor the null answer by a "$\theta$" threshold, where "$\theta$" was optimized based on dev set performance (see Section 4.3 in Devlin). We use purely the softmax probabilities output by the model without favoring no answer by an optimized threshold based on the dev set itself.

\subsection{Where to exact embeddings}
\label{apdx:where_to_extract_embeddings}

We wished to extract embeddings at two stages in the training process: 1. Early-on so that BERT fine-tuning is cheap, and the embeddings are amenable to use as data for modeling, 2. At a slightly later stage before convergence so that our models have a chance to achieve or outcompete the best performing BERT model. 

For span detection, our learning curve for full-epoch BERT fine-tuning shows that 1 epoch is relatively decent compared to peak performance at 2 epochs, which gives us a chance to outperform our best observed BERT performance (goal 2). At the same time, fine-tuning for a full epoch on all of SQuAD 2.0 takes around 5 hours, which is already very expensive. To this end, we explored BERT performance every 1/10 of a fractional epoch between 0 and 1. We see that both EM and F1 are increasing steadily in our single run, with a large jump between fractional epochs 3 and 4. In addition to 1 epoch, we extracted embeddings at 3/10 of an epoch immediately before this jump, which took approximately 1.5 hours of fine-tuning. We believe this is a decent spot because such partial fine-tuning is relatively cheap, and performance has yet to jump, giving us an opportunity to improve performance using our models (goal 1).

For classification, looking at our learning curve for full-epoch fine-tuning, 1 epoch is yet again a clear candidate for extracting the embeddings (goal 2). Performance is already relatively decent compared to peak performance at 3 epochs, but takes $\frac{1}{3}$ time and compute resources to train. For goal 1, we extracted embeddings at 2/10 of a fractional epoch between 0 and 1. We see a clear jump in performance between 1/10 of a fractional epoch and 2/10, a jump we do not observe again until 8/10. Therefore, from the dev set perspective, 2/10 is a high value fractional epoch, whether 3/10 - 7/10 does not add much to the performance of the model.

\subsection{Need to fine-tune}
\label{adpx:need_to_fine_tune}

Using a variety of parameter-efficient CNNs and dense architectures, we found such models were unable to fit BERT hidden state activations without fine-tuning to our specific QA task SQuAD 2.0. This discovery is consistent with the observations made in \cite{ma2019universal}, where the authors found that fine-tuned BERT on either SNLI and in-domain text corpus consistently outperformed pre-trained BERT without fine-tuning by a large margin for various tasks, including QA (although not on SQuAD). As a result, we decided mild amounts of fine-tuning was necessary in order to generate usable embeddings.

\subsection{Training hardware and development pipeline}
\label{apdx:training_hardware}

All span annotation training and inferencing was performed on a Microsoft Azure data science virtual machine with 112GiB onboard ram and a single NVIDIA Tesla TITAN series V100 Tensor Core GPU capable of ~7 TFLOPS double-precision and tensor performance of ~112 TFLOPS and possessed 16 GiB onboard RAM.  All binary classification training and experimentation was performed on an IBM virtual machine equipped with two NVIDIA Tesla TITAN series V100 Tensor Core GPU's (see [\ref{apdx:bertvision_development_pipeline_graph}]).

\begin{figure*}[t]
	\centering
	\includegraphics[width=\textwidth]{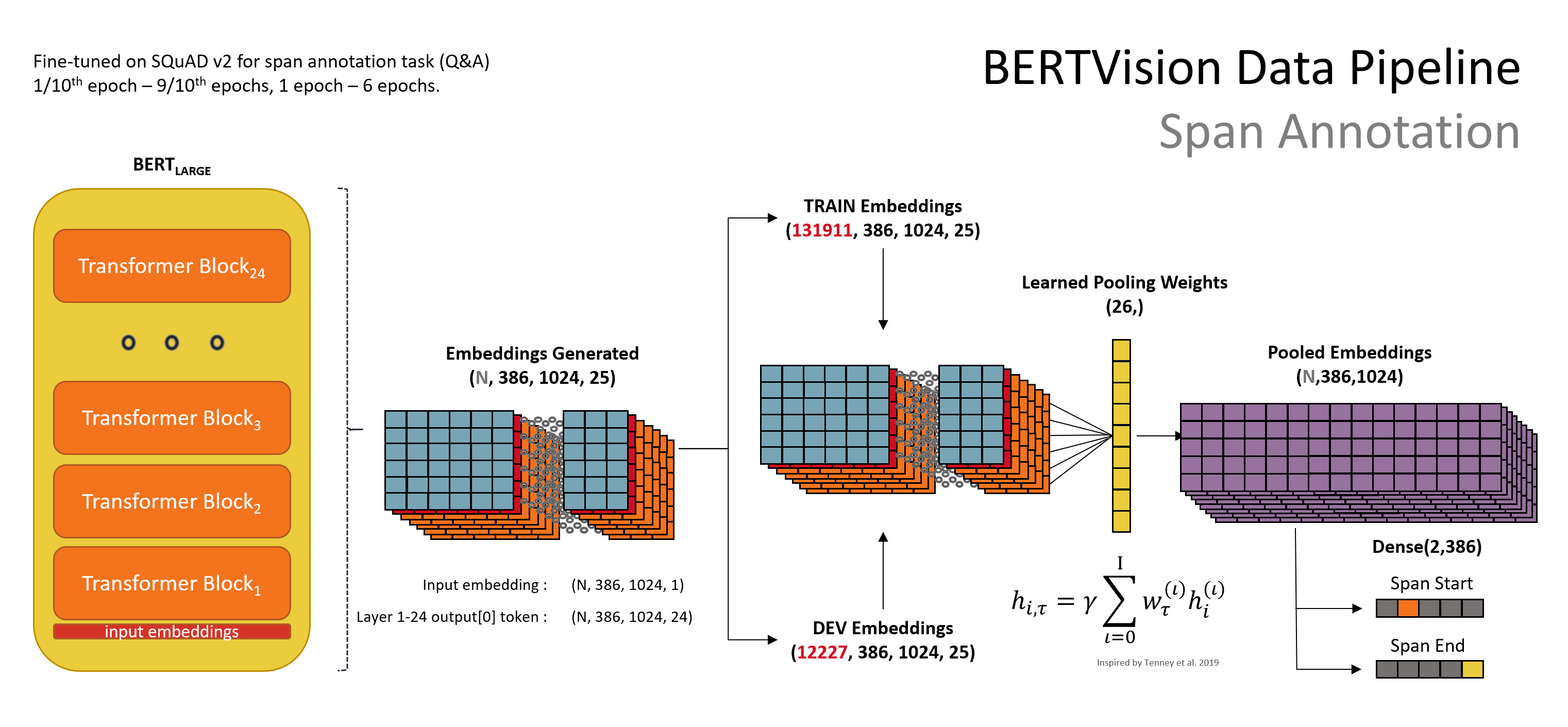}%
	\caption{BERTVision span annotation data pipeline}
	\label{apdx:bertvision_span_annotation_data_pipeline_graph}
\end{figure*}

\begin{figure*}[!h]
	\centering
	\includegraphics[width=\textwidth]{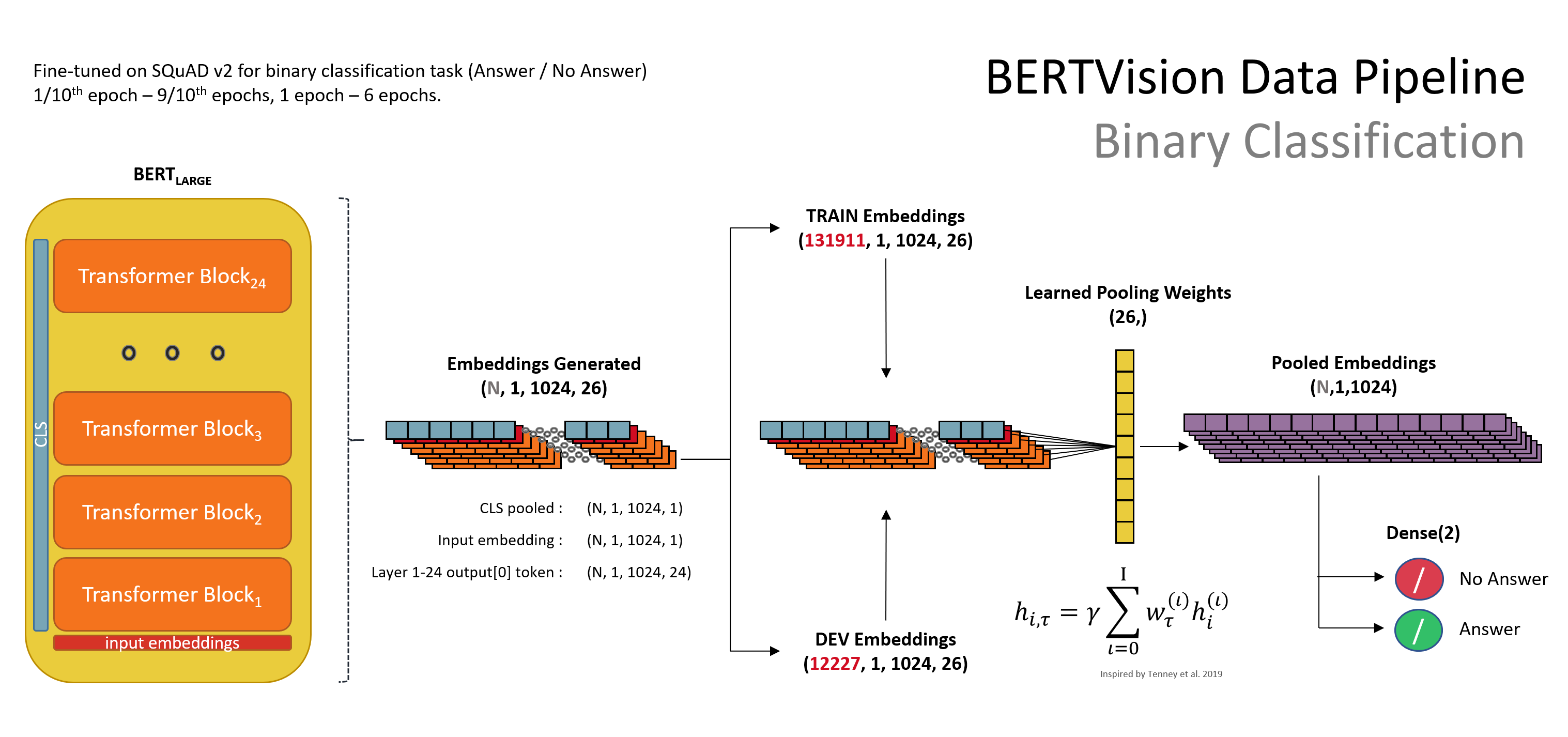}%
	\caption{BERTVision classification data pipeline}
	\label{apdx:bertvision_classification_data_pipeline_graph}
\end{figure*}

\newpage

\subsection{Non-uniform weights}
\label{apdx:non-uniform}

Learned weights for learning pooling using the approach described in \cite{tenney-etal-2019-bert}. The model favors using later layers especially after layer 17. This is consistent with the observation in \cite{van_Aken_2019} that the last layers of BERT-base are much more accurate at supporting fact identification (which is the author’s proxy for span identification).

\end{document}